\newif\ifCONF
\CONFfalse
\newif\ifarXiv
\arXivfalse
\newif\ifWP
\WPfalse
\newif\ifFULL
\FULLfalse

\arXivtrue
\newif\ifnotCONF	% derivative conditional (notCONF = arXiv or WP or FULL)
\notCONFtrue
\ifCONF\notCONFfalse\fi

\ifCONF
  \newcommand{\GTPVIII}{\cite{vovk/etal:2005AIStatslocal}}
  \newcommand{\GTPXIII}{\cite{GTP13local}}
  \newcommand{\GTPXIV}{\cite{GTP14local}}
\fi
\ifarXiv
  \newcommand{\GTPVIII}{\cite{GTP8arXiv}}
  \newcommand{\GTPXIII}{\cite{GTP13arXiv}}
  \newcommand{\GTPXIV}{\cite{GTP14arXiv}}
\fi
\ifWP
  \newcommand{\GTPVIII}{\cite{GTP8}}
  \newcommand{\GTPXIII}{\cite{GTP13}}
  \newcommand{\GTPXIV}{\cite{GTP14}}
\fi

\ifCONF
\documentclass{llncs}
\usepackage{amsmath,amsfonts,latexsym}
\newcommand{\Extra}[1]{}
\fi

\ifarXiv
\documentclass{article}
\usepackage{amsmath,amsfonts,latexsym}
\newcommand{\Extra}[1]{}
\fi

\ifWP
\documentclass[toc]{gtarticle}
\usepackage{amsmath,amsfonts,latexsym,epsfig}
\renewcommand{\Extra}[1]{#1}
\fi

\allowdisplaybreaks[3]

\newcommand{\Vladimir}{Vladimir }
\newcommand{\DOT}{.}

\newcommand{\st}{\mathop{|}}
\newcommand{\D}{\,\mathrm{d}}
\newcommand{\co}{\mathop{\overline{{\rm co}}}}
\newcommand{\diam}{\mathop{\rm diam}\nolimits}

\newcommand{\K}{\mathcal{K}}		% capital
\newcommand{\KKK}{\mathbf{k}}		% kernel
\newcommand{\CCC}{\mathbf{c}}		% constant
\newcommand{\III}{\mathbb{I}}
\newcommand{\FFF}{\mathcal{F}}		% function space
\newcommand{\SSS}{\mathcal{S}}		% Sobolev space

\ifnotCONF
\newcommand{\bbbr}{\mathbb{R}}
\newtheorem{lemma}{Lemma}

\newtheorem{theorem}{Theorem}
\newenvironment{proof}
  {\trivlist\item[\hskip\labelsep\textbf{Proof}]}
  {\endtrivlist}
\fi

\newcommand{\boxforqed}{\rule{.3em}{1.5ex}}
\newcommand{\qedtext}{\unskip\nobreak\hfil
  \penalty50\hskip1em\null\nobreak\hfil\boxforqed
  \parfillskip=0pt\finalhyphendemerits=0\endgraf}
\newcommand{\qedmath}{\tag*{\boxforqed}}
\newenvironment{remark*}
  {\trivlist\item[\hskip\labelsep{\bfseries Remark}]\relax}
  {\endtrivlist}

\newlength{\IndentI}
\newlength{\IndentII}
\newlength{\IndentIII}
\newlength{\WidthI}
\newlength{\WidthII}
\newlength{\WidthIII}
\ifnotCONF
  \newcommand{\UC}[1]{#1}
\fi
\ifCONF
  \newcommand{\UC}[1]{\uppercase{#1}}
\fi

\title{Defensive \UC{f}orecasting for \UC{l}inear \UC{p}rotocols}

\ifCONF		% for ALT'2005 camera ready
\author{Vladimir Vovk\inst{1} \and Ilia Nouretdinov\inst{1}
  \and Akimichi Takemura\inst{2} \and Glenn Shafer\inst{3,1}}

\institute{Department of Computer Science,
Royal Holloway, University of London,
Egham, Surrey TW20 0EX, England,
\email{{\rm\{}vovk,ilia,glenn{\rm\}@}cs.rhul.ac.uk}
\and
Department of Mathematical Informatics,
Graduate School of Information Science and Technology,
University of Tokyo,
7-3-1 Hongo, Bunkyo-ku, Tokyo 113-0033, Japan,
\email{takemura{\rm@}stat.t.u-tokyo.ac.jp}
\and
Rutgers Business School -- Newark and New Brunswick,
180 University Avenue, Newark, New Jersey 07102 USA,
\email{gshafer{\rm @}andromeda.rutgers.edu}}
\fi

\ifarXiv
\author{Vladimir Vovk\\
\texttt{vovk{\rm@}cs.rhul.ac.uk}\\
\texttt{http://vovk.net}
\and
Ilia Nouretdinov\\
\texttt{ilia{\rm@}cs.rhul.ac.uk}
\and
Akimichi Takemura\\
\texttt{takemura{\rm@}stat.t.u-tokyo.ac.jp}\\
\texttt{http://www.e.u-tokyo.ac.jp/\~{}takemura}
\and
Glenn Shafer\\
\texttt{gshafer{\rm@}andromeda.rutgers.edu}\\
\texttt{http://glennshafer.com}}
\fi

\ifWP
\author{Vladimir Vovk, Ilia Nouretdinov,\\
Akimichi Takemura, and Glenn Shafer}

\twodatestrue

\fi

\begin{document}
\maketitle
\begin{abstract}
% \looseness=-1
  We consider a general class of forecasting protocols,
  called ``linear protocols'',
  and discuss 
  several important special cases,
  including multi-class forecasting.
  Forecasting is formalized as a game between three players:
  Reality,
  whose role is to generate observations;
  Forecaster, whose goal is to predict the observations;
  and Skeptic, who tries to make money on any lack of agreement
  between Forecaster's predictions and the actual observations.
  Our main mathematical result
  is that for any continuous strategy for Skeptic in a linear protocol
  there exists a strategy for Forecaster
  that does not allow Skeptic's capital to grow.
  This result is a meta-theorem
  that allows one to transform any continuous law of probability
  in a linear protocol
  into a forecasting strategy whose predictions are guaranteed to satisfy this law.
  We apply this meta-theorem to a weak law of large numbers in Hilbert spaces
  to obtain a version of the K29 prediction algorithm for linear protocols
  and show that this version also satisfies the attractive properties
  of proper calibration and resolution
  under a suitable choice of its kernel parameter,
  with no assumptions about the way the data is generated.
\end{abstract}

\section{Introduction}
\label{sec:introduction}

\ifCONF
  In a recent paper,
  \GTPVIII,
\fi
\ifnotCONF
  In \GTPVIII{}
\fi
we suggested a new methodology for designing forecasting strategies.
Considering only the simplest case of binary forecasting,
we showed that any constructive,
in the sense explained below,
law of probability
can be translated into a forecasting strategy that satisfies this law.
In this paper this result is extended to a general class of protocols
including multi-class forecasting.
In proposing this approach to forecasting
we were inspired by \cite{foster/vohra:1998}
and papers further developing \cite{foster/vohra:1998},
although our methods and formal results appear to be completely different.
% and different aspects of our suggestion were independently, and somewhat earlier,
% proposed in \cite{kakade/foster:2004} and \cite{sandroni:2003}.

Whereas the meta-theorem stated in \GTPVIII{} is mathematically trivial,
the generalization considered in this paper is less so,
depending on the Schauder-Tikhonov fixed-point theorem.
Our general meta-theorem is stated in \S\ref{sec:meta-theorem}%
\ifnotCONF
  \ and proved in \S\ref{sec:meta-theorem} and Appendix \ref{app:meta-theorem}\fi.
The general forecasting protocols covered by this result
are introduced and discussed
in \S\S\ref{sec:forecasting}--\ref{sec:linear-protocol}.

In \GTPVIII{} we demonstrated the value of the meta-theorem
by applying it to the strong law of large numbers,
obtaining from it a kernel forecasting strategy which we called K29.
The derivation, however,
was informal,
involving heuristic transitions to a limit,
and this made it impossible to state formally any properties of K29.
In this paper we deduce K29 in a much more direct way from
the weak law of large numbers and state its properties.
(For binary forecasting,
this was also done in \GTPXIII,
and the reader might prefer to read that paper first.)
The weak law of large numbers is stated and proved in \S\ref{sec:WLLN},
and K29 is derived and studied in \S\ref{sec:K29}.

We call the approach to forecasting using our meta-theorem
``defensive forecasting'':
Forecaster is trying to defend himself when playing against Skeptic.
The justification of this approach given in this paper and in \GTPXIII{}
is K29's properties of proper calibration and resolution.
Another justification,
in a sense the ultimate justification of any forecasts,
is given in \GTPXIV:
%under suitable conditions,
defensive forecasts lead to good decisions;
this result, however, is obtained in \GTPXIV\ for rather simple decision problems
requiring only binary forecasts,
and its extensions will require this paper's results or their generalizations.

The exposition of probability theory
needed for this paper
is given in \cite{shafer/vovk:2001}.
The standard exposition is based on Kolmogorov's measure-theoretic axioms of probability,
whereas \cite{shafer/vovk:2001} states several key laws of probability
in terms of a game between the forecaster, the reality,
and a third player, the skeptic.
The game-theoretic laws of probability in \cite{shafer/vovk:2001}
are constructive
in that we explicitly construct computable winning strategies
for the forecaster in various games of forecasting.

\section{Forecasting as a \UC{g}ame}
\label{sec:forecasting}

Following \cite{shafer/vovk:2001} and \GTPVIII{}
we consider the following general forecasting protocol:

\bigskip

\noindent
\textsc{Forecasting Game 1}

\noindent
\textbf{Players:} Reality, Forecaster, Skeptic

\hangindent=\IndentI
\noindent
\textbf{Parameters:}
$\mathbf{X}$ (\emph{data space}),
$\mathbf{Y}$ (\emph{observation space}),
$\mathbf{F}$ (\emph{Forecaster's move space}),
$\mathbf{S}$ (\emph{Skeptic's move space}),
$\lambda:\mathbf{S}\times\mathbf{F}\times\mathbf{Y}\to\bbbr$
(\emph{Skeptic's gain function} and \emph{Forecaster's loss function})

\noindent
\textbf{Protocol:}

\parshape=8
\IndentI   \WidthI
\IndentI   \WidthI
\IndentII  \WidthII
\IndentII  \WidthII
\IndentII  \WidthII
\IndentII  \WidthII
\IndentII  \WidthII
\IndentI   \WidthI
\noindent
$\K_0 := 1$.\\
FOR $n=1,2,\ldots$:\\
  Reality announces $x_n\in\mathbf{X}$.\\
  Forecaster announces $f_n\in\mathbf{F}$.\\
  Skeptic announces $s_n\in\mathbf{S}$.\\
  Reality announces $y_n\in\mathbf{Y}$.\\
  $\K_n := \K_{n-1} + \lambda(s_n,f_n,y_n)$.\\
END FOR

\noindent
\textbf{Restriction on Skeptic:}
Skeptic must choose the $s_n$
so that his capital is always nonnegative
($\K_n \ge 0$ for all $n$)
no matter how the other players move.

\bigskip

\noindent
This is a perfect-information protocol:
the players move in the order indicated,
and each player sees the other player's moves
as they are made.
It specifies
both an initial value for Skeptic's capital ($\K_0 = 1$)
and a lower bound on its subsequent values ($\K_n \ge 0$).
We will say that $x_n$ are the \emph{data},
$y_n$ are the \emph{observations},
% $(x_n,y_n)$ are the \emph{examples},
and $f_n$ are the \emph{forecasts}.
In applications,
the datum $x_n$ will contain all available information deemed useful in forecasting $y_n$.
% Sometimes there will be a natural bijection
% between $\mathbf{Y}$ and another set $\mathbf{Y}'$
% (see, e.g., the bounded mean-variance forecasting protocol below),
% and we will then identify $y'\in\mathbf{Y}'$
% with the corresponding $y\in\mathbf{Y}$;
% in particular,
% the elements of $\mathbf{Y}'$ will also be called observations.

Book \cite{shafer/vovk:2001} contains several results
(game-theoretic versions of limit theorems of probability theory)
of the following form:
Skeptic has a strategy that guarantees
that either a property of agreement between the forecasts $f_n$
and observations $y_n$ is satisfied
or Skeptic becomes very rich
(without risking bankruptcy,
according to the protocol).
All specific strategies considered in \cite{shafer/vovk:2001}
have computable versions.
According to Brouwer's principle
(see, e.g., \S1 of \cite{stoltenberg-hansen/tucker:2003}
for a recent review of the relevant literature)
they must be automatically continuous;
in any case,
their continuity can be checked directly.
In \GTPVIII{} we showed that,
under a special choice of the players' move spaces
and Skeptic's gain function $\lambda$,
for any continuous strategy for Skeptic
Forecaster has a strategy that guarantees that Skeptic's capital
never increases when he plays that strategy.
Therefore,
Forecaster has strategies that ensure various properties
of agreement between the forecasts and the observations.

The purpose of this paper is to extend the result of \GTPVIII{}
to a wide class of Skeptic's gain functions $\lambda$.
But first we consider
several important special cases of Forecasting Game 1.

\subsection*{Binary \UC{f}orecasting}

The simplest non-trivial case,
considered in \GTPVIII,
is where $\mathbf{Y}=\{0,1\}$,
$\mathbf{F}=[0,1]$,
$\mathbf{S}=\bbbr$,
and
\begin{equation}\label{eq:gain}
  \lambda(s_n,f_n,y_n)
  =
  s_n(y_n-f_n)\ifCONF\;\fi.
\end{equation}
Intuitively,
Forecaster gives probability forecasts for $y_n$:
$f_n$ is his subjective probability that $y_n=1$.
The operational interpretation of $f_n$
is that it is the price that Forecaster charges for a ticket
that will pay $y_n$ at the end of the $n$th round of the game;
$s_n$ is the number (positive, zero, or negative)
of such tickets that Skeptic chooses to buy.

\subsection*{Bounded \UC{r}egression}

This is the most straightforward extension of binary forecasting,
considered in \cite{shafer/vovk:2001}, \S3.2.
The move spaces are
$\mathbf{Y}=\mathbf{F}=[A,B]$,
where $A$ and $B$ are two constants,
and $\mathbf{S}=\bbbr$;
the gain function is, as before, (\ref{eq:gain}).
This protocol allows one to prove a strong law of large numbers
(\cite{shafer/vovk:2001}, Proposition~3.3)
and a simple one-sided law of the iterated logarithm
(\cite{shafer/vovk:2001}, Corollary~5.1).

\subsection*{Multi-class \UC{f}orecasting}

Another extension of binary forecasting is the protocol where
$\mathbf{Y}$ is a finite set,
$\mathbf{F}$ is the set of all probability distributions on $\mathbf{Y}$,
$\mathbf{S}$ is the set of all real-valued functions on $\mathbf{Y}$,
and
\begin{equation*}
  \lambda(s_n,f_n,y_n)
  =
  s_n(y_n)
  -
  \int s_n \D f_n
  \ifCONF\;\fi.
\end{equation*}
The intuition behind Skeptic's move $s_n$ is that Skeptic
buys the ticket which pays $s_n(y_n)$ after $y_n$ is announced;
he is charged $\int s_n \D f_n$ for this ticket.

The binary forecasting protocol is ``isomorphic''
to the special case of this protocol
where $\mathbf{Y}=\{0,1\}$:
Forecaster's move $f_n$ in the binary forecasting protocol
is represented by the probability distribution $f'_n$ on $\{0,1\}$
assigning weight $f_n$ to $\{1\}$
and Skeptic's move $s_n$ in the binary forecasting protocol
is represented by any function $s'_n$ on $\{0,1\}$
such that $s'_n(1)-s'_n(0)=s_n$.
The isomorphism between these two protocols follows from
\begin{multline*}
  s'_n(y_n)
  -
  \int s'_n \D f'_n
  =
  s'_n(y_n)
  -
  s'_n(1) f_n
  -
  s'_n(0) (1-f_n)\\
  =
  s'_n(y_n) - s'_n(0)
  -
  s_n f_n
  =
  s_n(y_n-f_n)
\end{multline*}
(remember that $y_n\in\{0,1\}$).  

\ifFULL
Sometimes Skeptic is allowed to buy only non-negative amount of tickets,
i.e.\ $\mathbf{S}$ is the set of all non-negative functions on $\mathbf{Y}$.
In this case, Forecaster can ask higher prices for the tickets
and his move is not necessarily a probability distribution on $\mathbf{Y}$.
We will clarify this situation in Section \ref{sec:linear-protocol}.
\fi

\subsection*{Bounded \UC{m}ean-\UC{v}ariance \UC{f}orecasting}

In this protocol,
$\mathbf{Y}
  =
  [A,B]
$,
where $A$ and $B$ are again two constants,
$\mathbf{F}
  =
  \mathbf{S}
  =
  \bbbr^2
$,
and
\begin{equation*}
  \lambda(s_n,f_n,y_n)
  =
  \lambda((M_n,V_n),(m_n,v_n),y_n)
  =
  M_n(y_n-m_n)
  +
  V_n((y_n-m_n)^2-v_n)
  \ifCONF\;\fi.
\end{equation*}
Intuitively,
Forecaster is asked to forecast $y_n$
with a number $m_n$ and also forecast the accuracy $(y_n-m_n)^2$
of his first forecast with a number $v_n$.
This protocol, although usually without the restriction $y_n\in[A,B]$,
is used extensively in \cite{shafer/vovk:2001}
(e.g., in Chaps.~4 and~5).

An equivalent representation of this protocol is
$
  \mathbf{Y}
  =
  \{
    (t,t^2)
    \st
    t\in[A,B]
  \}
$,
$
  \mathbf{F}
  =
  \mathbf{S}
  =
  \bbbr^2
$
and
\begin{equation*}
  \lambda(s_n,f_n,y_n)
  =
  \lambda((s'_n,s''_n),(f'_n,f''_n),(t_n,t^2_n))
  =
  s'_n(t_n-f'_n)
  +
  s''_n(t^2_n-f''_n)
  \ifCONF\;\fi.
\end{equation*}
% The intuition is that Forecaster is asked
% not only to forecast the observation $t_n$ to be chosen by Reality
% with a number $f'_n$
% but also to estimate the square $t^2_n$ of the observation with a number $f''_n$.
The equivalence of the two representations can be seen as follows:
Reality's move $(x_n,t_n)$ in the first representation
corresponds to $(x_n,y_n)=(x_n,(t_n,t_n^2))$ in the second representation,
Forecaster's move $(m_n,v_n)$ in the first representation
corresponds to $(f'_n,f''_n)=(m_n,v_n+m_n^2)$ in the second representation,
and Skeptic's move $(s'_n,s''_n)$ in the second representation
corresponds to $(M_n,V_n)=(s'_n+2m_ns''_n,s''_n)$ in the first representation.
This establishes a bijection between Reality's move spaces,
a bijection between Forecaster's move spaces,
and a bijection between Skeptic's move spaces
in the two representations;
Skeptic's gains are also the same in the two representations:
\begin{multline*}
  s'_n(t_n-f'_n)
  +
  s''_n(t_n^2-f''_n)\\
  =
  s'_n(t_n-m_n)
  +
  s''_n
  \Bigl(
    \bigl(
      (t_n-m_n)^2 + 2(t_n-m_n)m_n + m_n^2
    \bigr)
    -
    \bigl(
      v_n+m_n^2
    \bigr)
  \Bigr)\\
  =
  (s'_n+2m_ns''_n)(t_n-m_n)
  +
  s''_n
  \bigl(
    (t_n-m_n)^2 - v_n
  \bigr)
  \ifCONF\;\fi.
\end{multline*}

\section{Linear \UC{p}rotocol}
\label{sec:linear-protocol}

Forecasting Game 1 is too general to derive results of the kind we are interested in.
In this subsection we will introduce a narrower protocol
which will still be wide enough to cover all special cases considered so far.

All move spaces are now subsets of a Hilbert space $\mathbf{L}$
(we allow $\mathbf{L}$ to be non-separable or finite-dimensional;
in fact, in this paper we emphasize the case where $\mathbf{L}=\bbbr^m$ for some positive integer $m$).
The observation space is a non-empty pre-compact subset $\mathbf{Y}\subset\mathbf{L}$
(we say that a set is \emph{pre-compact} if its closure is compact;
if $\mathbf{L}=\bbbr^m$,
this is equivalent to it being bounded),
Forecaster's move space $\mathbf{F}$ is the whole of $\mathbf{L}$,
and Skeptic's move space $\mathbf{S}$ is also the whole of $\mathbf{L}$.
Skeptic's gain function is
\begin{equation*}
  \lambda(s_n,f_n,y_n)
  =
  \left\langle
    s_n,
    y_n-f_n
  \right\rangle_{\mathbf{L}}
  \ifCONF\;\fi.
\end{equation*}
Therefore,
we consider the following perfect-information game:

\bigskip

\noindent
\textsc{Forecasting Game 2}

\noindent
\textbf{Players:} Reality, Forecaster, Skeptic

\hangindent=\IndentI
\noindent
\textbf{Parameters:}
$\mathbf{X}$,
$\mathbf{L}$ (Hilbert space),
$\mathbf{Y}$ (non-empty pre-compact subset of $\mathbf{L}$)

\noindent
\textbf{Protocol:}

\parshape=8
\IndentI   \WidthI
\IndentI   \WidthI
\IndentII  \WidthII
\IndentII  \WidthII
\IndentII  \WidthII
\IndentII  \WidthII
\IndentII  \WidthII
\IndentI   \WidthI
\noindent
$\K_0:=1$.\\
FOR $n=1,2,\ldots$:\\
  Reality announces $x_n\in\mathbf{X}$.\\
  Forecaster announces $f_n\in \mathbf{L}$.\\
  Skeptic announces $s_n\in\mathbf{L}$.\\
  Reality announces $y_n\in\mathbf{Y}$.\\
  \hbox to \WidthII
  {$\K_n := \K_{n-1} + \langle s_n, y_n-f_n\rangle_{\mathbf{L}}$.
  \refstepcounter{equation}\label{eq:K}
  \hfil (\theequation)}\\
END FOR

\noindent
\textbf{Restriction on Skeptic:}
Skeptic must choose the $s_n$
so that his capital is always nonnegative
no matter how the other players move.

\bigskip

Let us check that the specific protocols considered
in the previous section are covered by this \emph{linear protocol}
(and for all those protocols $\mathbf{L}$ can be taken finite dimensional,
$\mathbf{L}=\bbbr^m$ for some $m\in\{1,2,\ldots\}$).
At first sight, even the binary forecasting protocol is not covered,
as Forecaster's move space is $\mathbf{F}=[0,1]$ rather than $\bbbr$.
It is easy to see, however, that Forecaster's move $f_n\notin\co\mathbf{Y}$
outside the convex closure $\co\mathbf{Y}$ of the observation space
(the convex closure $\co A$ of a set $A$ is defined to be the intersection
of all convex closed sets containing $A$)
is always inadmissible,
in the sense that there exists Skeptic's reply $s_n$ making him arbitrarily rich
regardless of Reality's move,
and so we can as well choose $\mathbf{F}:=\co\mathbf{Y}$.
Indeed,
suppose that $f_n \notin \co\mathbf{Y}$ in the linear protocol.
Since $\mathbf{Y}$ is pre-compact,
$\co\mathbf{Y}$ is compact
(\cite{rudin:1991}, Theorem 3.20(c)).
By the Hahn-Banach theorem
(\cite{rudin:1991}, Theorem 3.4(b)),
there exists a vector $s_n\in\mathbf{L}$ such that
\begin{equation*}
  \inf_{y\in\mathbf{Y}}
  \left\langle
    s_n,
    y - f_n
  \right\rangle_{\mathbf{L}}
  > 0
  \ifCONF\;\fi.
\end{equation*}
(It would have been sufficient for either $\{f_n\}$ or $\co\mathbf{Y}$ to be compact;
in fact both are.)
Skeptic's move $Cs_n$ can make him as rich as he wishes
as $C$ can be arbitrarily large.
In what follows,
we will usually assume that Forecaster's move space is $\co\mathbf{Y}$
and use $\mathbf{F}$ as a shorthand for $\co\mathbf{Y}$.

Now it is obvious that
the binary forecasting, bounded regression,
and bounded mean-variance forecasting (in its second representation) protocols
are special cases
of the linear protocol
(perhaps with $\mathbf{F}=\co\mathbf{Y}$).
For the multi-class forecasting protocol,
we should represent $\mathbf{Y}$ as the vertices
\begin{equation*}
  y^1:=(1,0,0,\ldots,0),
  \enspace
  y^2:=(0,1,0,\ldots,0),\ldots,
  \enspace
  y^m:=(0,0,0,\ldots,1)
\end{equation*}
of the standard simplex in $\bbbr^m$,
where $m$ is the size of $\mathbf{Y}$,
represent the probability distributions $f$ on $\mathbf{Y}$
as vectors $(f\{y^1\},\ldots,f\{y^m\})$ in $\bbbr^m$,
and represent the real-valued functions $s$ on $\mathbf{Y}$
as vectors $(s(y^1),\ldots,s(y^m))$ in $\bbbr^m$.
% If Skeptic is allowed to buy only non-negative amount of tickets
% his move space is the positive orthant $\mathbf{S} = \bbbr_+^m = [0,\infty)^m$.

\section{Meta-theorem}
\label{sec:meta-theorem}

In this section we state the main mathematical result of this paper:
for any continuous strategy for Skeptic
there exists a strategy for Forecaster
that does not allow Skeptic's capital to grow,
regardless of what Reality is doing.
As in \GTPVIII,
we make Skeptic announce his strategy for each round
at the outset of that round
rather than announce his strategy for the whole game
at the beginning of the game,
and we drop all restrictions on Skeptic.
Forecaster's move space is restricted to $\mathbf{F} = \co\mathbf{Y}$. 
The resulting perfect-information game is:

\bigskip

\noindent
\textsc{Forecasting Game 3}

\noindent
\textbf{Players:} Reality, Forecaster, Skeptic

\hangindent=\IndentI
\noindent
\textbf{Parameters:}
$\mathbf{X}$,
$\mathbf{L}$ (Hilbert space),
$\mathbf{Y}\subset\mathbf{L}$ (non-empty and pre-compact)

\noindent
\textbf{Protocol:}

\parshape=8
\IndentI   \WidthI
\IndentI   \WidthI
\IndentII  \WidthII
\IndentII  \WidthII
\IndentII  \WidthII
\IndentII  \WidthII
\IndentII  \WidthII
\IndentI   \WidthI
\noindent
$\K_0$ is set to a real number.\\
FOR $n=1,2,\ldots$:\\
  Reality announces $x_n\in\mathbf{X}$.\\
  Skeptic announces continuous $S_n:\co\mathbf{Y}\to\mathbf{L}$.\\
  Forecaster announces $f_n\in\co\mathbf{Y}$.\\
  Reality announces $y_n\in\mathbf{Y}$.\\
  $\K_n := \K_{n-1} + \langle S_n(f_n), y_n-f_n\rangle_{\mathbf{L}}$.\\
END FOR

\bigskip

\begin{theorem}\label{thm:main}
  Forecaster has a strategy in Forecasting Game 3
  that ensures $\K_0\ge\K_1\ge\K_2\ge\cdots$.
\end{theorem}
\begin{proof}
  Fix a round $n$ and Skeptic's move $S_n:\mathbf{F}\to\mathbf{L}$
  (we will refer to $S_n$ as a vector field in $\mathbf{F}$).
  Our task is to prove the existence of a point $f_n\in\mathbf{F}$
  such that, for all $y\in\mathbf{Y}$,
  $\langle S_n(f_n),y-f_n\rangle_{\mathbf{L}}\le0$.

  If for some $f\in\partial\mathbf{F}$
  (we use $\partial A$ to denote the boundary of $A\subseteq\mathbf{L}$)
  the vector $S_n(f)$
  is normal and directed exteriorly to $\mathbf{F}$
  (in the sense that $\langle S_n(f),y-f\rangle_{\mathbf{L}}\le0$
  for all $y\in\mathbf{F}$),
  we can take such $f$ as $f_n$.
  Therefore,
  we assume, without loss of generality,
  that $S_n$ is never normal and directed exteriorly on $\partial\mathbf{F}$.
  Then by Lemma \ref{lem:fixed-point} in Appendix \ref{app:meta-theorem}
  there exists $f$ such that $S_n(f)=0$,
  and we can take such $f$ as $f_n$.
  \qedtext
\end{proof}

\begin{remark*}
  Notice that Theorem~\ref{thm:main} will not become weaker
  if the first move by Reality
  (choosing $x_n$) is removed from each round of the protocol.
% As explained earlier,
% we can require that $f_n\in\co\mathbf{Y}$ for all $n$.
\end{remark*}

\section{A \UC{w}eak \UC{l}aw of \UC{l}arge \UC{n}umbers in \UC{H}ilbert \UC{s}pace}
\label{sec:WLLN}

Unfortunately, the usual law of large numbers
is not useful for the purpose of designing forecasting strategies
(see the discussion in \GTPVIII).
Therefore, we state a generalized law of large numbers;
at the end of this section
we will explain connections with the usual law of large numbers.
In this section we consider Forecasting Game 2
without the requirement $\K_0=1$
and with the restriction on Skeptic dropped.
If we fix a strategy for Skeptic
and Skeptic's initial capital $\K_0$
(not necessarily $1$ or even a positive number),
$\K_n$ defined by (\ref{eq:K})
becomes a function of Reality's and Forecaster's moves.
Such functions will be called \emph{capital processes}.

Let $\Phi:\mathbf{F}\times\mathbf{X}\to\mathbf{H}$
(as usual, $\mathbf{F}=\co\mathbf{Y}$)
be a \emph{feature mapping} into a Hilbert space $\mathbf{H}$;
$\mathbf{H}$ is called the \emph{feature space}.
The next theorem uses the notion of tensor product;
for details, see Appendix \ref{app:tensor}.
\begin{theorem}\label{thm:WLLN}
  The function
  \begin{equation}\label{eq:capital}
    \K_n
    :=
    \left\|
      \sum_{i=1}^n
      (y_i-f_i)
      \otimes
      \Phi(f_i,x_i)
    \right\|^2_{\mathbf{L}\otimes\mathbf{H}}
    -
    \sum_{i=1}^n
    \left\|
      y_i-f_i
    \right\|^2_{\mathbf{L}}
    \left\|
      \Phi(f_i,x_i)
    \right\|^2_{\mathbf{H}}
  \end{equation}
  is a capital process (not necessarily non-negative)
  of some strategy for Skeptic.
\end{theorem}
\begin{proof}
  We start by noticing that
  \begin{align*}
    \K_n
    -
    \K_{n-1}
    &=
    \left\|
      \sum_{i=1}^{n-1}
      (y_i-f_i)
      \otimes
      \Phi(f_i,x_i)
      +
      (y_n-f_n)
      \otimes
      \Phi(f_n,x_n)
    \right\|^2_{\mathbf{L}\otimes\mathbf{H}}\\
    &\quad{}-
    \left\|
      \sum_{i=1}^{n-1}
      (y_i-f_i)
      \otimes
      \Phi(f_i,x_i)
    \right\|^2_{\mathbf{L}\otimes\mathbf{H}}
    -
    \left\|
      y_n-f_n
    \right\|^2_{\mathbf{L}}
    \left\|
      \Phi(f_n,x_n)
    \right\|^2_{\mathbf{H}}\\
    &=
    2
    \left\langle
      \sum_{i=1}^{n-1}
      (y_i-f_i)
      \otimes
      \Phi(f_i,x_i),
      (y_n-f_n)
      \otimes
      \Phi(f_n,x_n)
    \right\rangle_{\mathbf{L}\otimes\mathbf{H}}\\
    &=
    2
    \sum_{i=1}^{n-1}
    \left\langle
      y_i-f_i,
      y_n-f_n
    \right\rangle_{\mathbf{L}}
    \left\langle
      \Phi(f_i,x_i),
      \Phi(f_n,x_n)
    \right\rangle_{\mathbf{H}}
  \end{align*}
  (in the last two equalities we used (\ref{eq:def2}) and (\ref{eq:implication})
  from Appendix \ref{app:tensor}).
  Introducing the notation
  \begin{equation}\label{eq:kernel1a}
    \KKK((f,x),(f',x'))
    :=
    \langle\Phi(f,x),\Phi(f',x')\rangle_{\mathbf{H}}\ifCONF\;\fi,
  \end{equation}
  where $(f,x),(f',x')\in\mathbf{F}\times\mathbf{X}$,
  we can rewrite the expression for $\K_n-\K_{n-1}$ as
  \begin{equation*}
    \left\langle
      2
      \sum_{i=1}^{n-1}
      \KKK((f_i,x_i),(f_n,x_n))
      (y_i-f_i),
      y_n-f_n
    \right\rangle_{\mathbf{L}}
    \ifCONF\;\fi.
  \end{equation*}
  Therefore, $\K_n$ is the capital process corresponding to Skeptic's strategy
  \begin{equation}\label{eq:strategy}
    2
    \sum_{i=1}^{n-1}
    \KKK((f_i,x_i),(f_n,x_n))
    (y_i-f_i)
    \ifCONF\;\fi;
  \end{equation}
  this completes the proof.
  \qedtext
\end{proof}

\subsection*{More \UC{s}tandard \UC{s}tatements of the \UC{w}eak \UC{l}aw}

In the rest of this section we explain connections of Theorem \ref{thm:WLLN}
with more standard statements of the weak law of large numbers;
in this part of the paper we will use some notions
introduced in \cite{shafer/vovk:2001}.
The rest of the paper does not depend on this material,
and the reader may wish to skip this subsection.
%go to p.~\pageref{sec:K29} now.

Let us assume that
\begin{equation*}
  \CCC_{\Phi}
  :=
  \sup_{(f,x)\in\mathbf{F}\times\mathbf{X}}
  \left\|
    \Phi(f,x)
  \right\|_{\mathbf{H}}
  <
  \infty\ifCONF\;\fi.
\end{equation*}
We will use the notation
$
  \diam(\mathbf{Y})
  :=
  \sup_{y,y'\in\mathbf{Y}}
  \left\|
    y-y'
  \right\|_{\mathbf{L}}
$;
it is clear that
$\diam(\mathbf{Y})<\infty$.
For any initial capital $\K_0$,
\begin{equation*}
  \K_n
  :=
  \K_0
  +
  \left\|
    \sum_{i=1}^n
    (y_i-f_i)
    \otimes
    \Phi(f_i,x_i)
  \right\|^2_{\mathbf{L}\otimes\mathbf{H}}
  -
  \sum_{i=1}^n
  \|y_i-f_i\|^2_{\mathbf{L}}
  \|\Phi(f_i,x_i)\|^2_{\mathbf{H}}
\end{equation*}
is the capital process of some strategy for Skeptic.
Suppose a positive integer $N$
(the duration of the game, or the \emph{horizon})
is given in advance
and $\K_0:=\diam^2(\mathbf{Y})\CCC_{\Phi}^2N$.
Then, in the game lasting $N$ rounds,
$\K_n$ is never negative and
\begin{equation*}
  \K_N
  \ge
  \left\|
    \sum_{i=1}^N
    (y_i-f_i)
    \otimes
    \Phi(f_i,x_i)
  \right\|^2_{\mathbf{L}\otimes\mathbf{H}}
  \ifCONF\;\fi.
\end{equation*}
If we do not believe that Skeptic can increase his capital $1/\delta$-fold
for a small $\delta>0$ without risking bankruptcy,
we should believe that
\begin{equation*}
  \left\|
    \sum_{i=1}^N
    (y_i-f_i)
    \otimes
    \Phi(f_i,x_i)
  \right\|^2_{\mathbf{L}\otimes\mathbf{H}}
  \le
  \diam^2(\mathbf{Y})\CCC_{\Phi}^2N/\delta
  \ifCONF\;\fi,
\end{equation*}
which can be rewritten as
\begin{equation}\label{eq:Bernoulli}
  \left\|
    \frac1N
%   \left(
      \sum_{i=1}^N
      (y_i-f_i)
      \otimes
      \Phi(f_i,x_i)
%   \right)
  \right\|_{\mathbf{L}\otimes\mathbf{H}}
  \le
  \diam(\mathbf{Y})
  \CCC_{\Phi}(N\delta)^{-1/2}\ifCONF\;\fi.
\end{equation}
In the terminology of \cite{shafer/vovk:2001},
\emph{the game-theoretic lower probability of the event (\ref{eq:Bernoulli})
is at least $1-\delta$}.

The game-theoretic version of Bernoulli's law of large numbers
is a special case of (\ref{eq:Bernoulli})
corresponding to $\Phi(f,x)=1$, for all $f$ and $x$,
$\mathbf{Y}=\{0,1\}$,
and $|\mathbf{X}|=1$
(the last two conditions mean that we are considering the binary forecasting protocol
without the data);
as usual, we assume that $f_i$ are chosen from $\co\mathbf{Y}=[0,1]$.
As explained in \cite{shafer/vovk:2001},
in combination with the measurability of Skeptic's strategy
guaranteeing (\ref{eq:Bernoulli}),
this implies that the measure-theoretic probability of the event (\ref{eq:Bernoulli})
is at least $1-\delta$,
assuming that the $y_i$ are generated by a probability distribution
and that each $f_i$ is the conditional probability that $y_i=1$
given $y_1,\ldots,y_{i-1}$.
This measure-theoretic result
was proved by Kolmogorov in 1929
(see \cite{kolmogorov:1929LLN})
and is the origin of the name ``K29 strategy''.

We will see in the next section
that the feature-space version (\ref{eq:Bernoulli})
of the weak law of large numbers
is much more useful than the standard version
for the purpose of forecasting.
% and it will turn out that K29 guarantees (\ref{eq:Bernoulli})
% with $\delta=1$.

\section{The K29 \UC{s}trategy and \UC{i}ts \UC{p}roperties}
\label{sec:K29}

According to Theorem~\ref{thm:main},
under the continuity assumption
there is a strategy for Forecaster that does not allow $\K_n$ to grow,
where $\K_n$ is defined by (\ref{eq:capital}).
Fortunately (but not unusually),
this strategy depends on the feature mapping $\Phi$
only via the corresponding \emph{kernel} $\KKK$
defined by (\ref{eq:kernel1a}).
The continuity assumption needed
is that $\KKK((f,x),(f',x'))$ should be continuous in $f$;
such kernels will be called \emph{admissible}.
According to (\ref{eq:strategy}),
the corresponding forecasting strategy,
which we will call the \emph{K29 strategy} with parameter $\KKK$,
is to output, on the $n$th round,
a forecast $f_n$ satisfying
\begin{equation*}
  S(f_n)
  :=
  \sum_{i=1}^{n-1}
  \KKK((f_i,x_i),(f_n,x_n))
  (y_i-f_i)
  =
  0
\end{equation*}
(or, if such $f_n$ does not exist,
the forecast is chosen to be a point $f_n\in\partial\mathbf{F}$
where $S(f_n)$ is normal and directed exteriorly to $\mathbf{F}$).

The protocol of this section is essentially that of Forecasting Game 3;
as Skeptic ceases to be an active player,
it simplifies to:

\medskip

\parshape=5
\IndentI   \WidthI
\IndentII  \WidthII
\IndentII  \WidthII
\IndentII  \WidthII
\IndentI   \WidthI
\noindent
FOR $n=1,2,\ldots$:\\
  Reality announces $x_n\in\mathbf{X}$.\\
  Forecaster announces $f_n\in\co\mathbf{Y}$.\\
  Reality announces $y_n\in\mathbf{Y}$.\\
END FOR

\medskip

\begin{theorem}\label{thm:K29}
%  In Forecasting Game 2,
  The K29 strategy guarantees that always
  \begin{equation}\label{eq:K29}
    \left\|
      \sum_{i=1}^n
      (y_i-f_i)
      \otimes
      \Phi(f_i,x_i)
    \right\|_{\mathbf{L}\otimes\mathbf{H}}
    \le
    \diam(\mathbf{Y})
    \CCC_{\Phi}\sqrt{n}\ifCONF\;\fi,
  \end{equation}
  where
  $\CCC_{\Phi}:=
  \sup_{(f,x)\in\mathbf{F}\times\mathbf{X}}
  \left\|
    \Phi(f,x)
  \right\|_{\mathbf{H}}$
  is assumed to be finite.
\end{theorem}
\begin{proof}
  The K29 strategy ensures that (\ref{eq:capital}) never increases;
  therefore,
% we always have
  \begin{equation*}
    \left\|
      \sum_{i=1}^n
      (y_i-f_i)
      \otimes
      \Phi(f_i,x_i)
    \right\|^2_{\mathbf{L}\otimes\mathbf{H}}
    \le
    \sum_{i=1}^n
    \left\|
      y_i-f_i
    \right\|^2_{\mathbf{L}}
    \left\|
      \Phi(f_i,x_i)
    \right\|^2_{\mathbf{H}}
    \le
    \diam^2(\mathbf{Y}) \CCC_{\Phi}^2 n
    \ifCONF\;\fi.
    \qedmath
  \end{equation*}
\end{proof}
\begin{remark*}
  The property (\ref{eq:K29}) is a special case of (\ref{eq:Bernoulli})
  corresponding to $\delta=1$;
  we gave an independent derivation
% for two reasons:
  to make our exposition self-contained
% and not to rely on the notions introduced in \cite{shafer/vovk:2001},
  and to avoid the extra assumptions used in the derivation of (\ref{eq:Bernoulli}),
  such as the horizon being finite and known in advance.
\end{remark*}

\subsection*{K29 with \UC{r}eproducing \UC{k}ernel Hilbert \UC{s}paces}

A \emph{reproducing kernel Hilbert space} (usually abbreviated to RKHS)
is a function space $\FFF$ on some set $Z$
such that all evaluation functionals $F\in\FFF\mapsto F(z)$, $z\in Z$,
are continuous.
We will be interested in RKHS on the Cartesian product $\mathbf{F}\times\mathbf{X}$.

By the Riesz-Fischer theorem,
for each $z\in Z$ there exists a function $\KKK_z\in\FFF$ such that
\begin{equation*}
  F(z)
  =
  \langle \KKK_z,F\rangle_{\FFF},
  \quad
  \forall F\in\FFF.
\end{equation*}
Let
\begin{equation}\label{eq:C}
  \CCC_{\FFF}
  :=
  \sup_{z\in Z}
  \left\|\KKK_z\right\|_{\FFF};
\end{equation}
we will be interested in the case $\CCC_{\FFF}<\infty$.

The \emph{kernel} of an RKHS $\FFF$ on $Z$ is
\begin{equation}\label{eq:kernel2}
  \KKK(z,z')
  :=
  \left\langle
    \KKK_z,\KKK_{z'}
  \right\rangle_{\FFF}
\end{equation}
(equivalently, we could define $\KKK(z,z')$ as $\KKK_z(z')$
or as $\KKK_{z'}(z)$).
It is clear that (\ref{eq:kernel2}) is a special case of the generalization
\begin{equation}\label{eq:kernel1b}
  \KKK(z,z')
  :=
  \langle\Phi(z),\Phi(z')\rangle_{\mathbf{H}}
\end{equation}
of (\ref{eq:kernel1a}).
In fact, the functions $\KKK$ that can be represented as (\ref{eq:kernel1b})
are exactly the functions that can be represented as (\ref{eq:kernel2});
they can be equivalently defined as symmetric positive definite
functions on $Z^2$
(see \GTPXIII{} for a list of references).

A long list of RKHS together with their kernels is given
in \cite{berlinet/thomas-agnan:2004}, \S7.4.
We will only give one example:
the Sobolev space $\SSS$
of absolutely continuous functions $F$ on $\bbbr$ with finite norm
\begin{equation}\label{eq:Sobolev}
  \left\|F\right\|_{\SSS}
  :=
  \sqrt
  {
    \int_{-\infty}^{\infty}
      F^2(z)
    \D z
    +
    \int_{-\infty}^{\infty}
      (F'(z))^2
    \D z
  };
\end{equation}
its kernel is
\begin{equation*}
  \KKK(z,z')
  =
  \frac{1}{2}
  \exp
  \left(
    -
    \left|
      z-z'
    \right|
  \right)
\end{equation*}
(see \cite{thomas-agnan:1996}
or \cite{berlinet/thomas-agnan:2004}, \S7.4, Example 24).
From the last equation we can see that $\CCC_{\SSS}=1/\sqrt{2}$.

The following is an easy corollary of Theorem~\ref{thm:K29}.
\begin{theorem}\label{thm:RKHS}
  Let $\FFF$ be an RKHS on $\mathbf{F}\times\mathbf{X}$.
% In Forecasting Game 2,
  The K29 strategy with parameter $\KKK$
  (defined by (\ref{eq:kernel2})) ensures
  \begin{equation}\label{eq:calibration-resolution}
    \left\|
      \sum_{i=1}^n
      F(f_i,x_i)
      (y_i-f_i)
    \right\|_{\mathbf{L}}
    \le
    \diam(\mathbf{Y})
    \CCC_{\FFF}
    \left\|
      F
    \right\|_{\FFF}
    \sqrt{n}
  \end{equation}
  for each function $F\in\FFF$,
  where $\CCC_{\FFF}$ is defined by (\ref{eq:C}).
\end{theorem}
\begin{proof}
  Let $\Phi:\mathbf{F}\times\mathbf{X}\to\mathbf{H}:=\FFF$ be defined by $\Phi(z):=\KKK_z$.
  Theorem~\ref{thm:K29}
% and the Cauchy-Schwarz inequality
% (in the form of Lemma \ref{lem:tensor2})
  then implies
  \begin{align*}
    \left\|
      \sum_{i=1}^n
      F(f_i,x_i)
      (y_i-f_i)
    \right\|_{\mathbf{L}}
    &=
    \left\|
      \sum_{i=1}^n
      \left\langle
        \KKK_{f_i,x_i},F
      \right\rangle_{\mathbf{H}}
      (y_i-f_i)
    \right\|_{\mathbf{L}}\\
    &=
    \left\|
      \sum_{i=1}^n
      \bigl(
        (y_i-f_i)\otimes\KKK_{f_i,x_i}
      \bigr)
      F
    \right\|_{\mathbf{L}}\\
    &\le
    \left\|
      \sum_{i=1}^n
      (y_i-f_i)\otimes\KKK_{f_i,x_i}
    \right\|_{\mathbf{L}\otimes\FFF}
    \left\|
      F
    \right\|_{\FFF}\\
    &\le
    \diam(\mathbf{Y})
    \CCC_{\FFF}
    \left\|F\right\|_{\FFF}
    \sqrt{n}
  \end{align*}
  (the second equality follows from Lemma~\ref{lem:tensor3}
  and the first inequality from Lemma~\ref{lem:tensor2} in Appendix \ref{app:tensor}).
% the inequality between the extreme terms of this chain
% is exactly (\ref{eq:calibration-resolution}).
  \qedtext
\end{proof}

\subsection*{Calibration and \UC{r}esolution}

Two important properties of a forecasting strategy
are its calibration and resolution,
which we introduce informally.
Our discussion in this section extends the discussion in \GTPXIII, \S5,
to the case of linear protocols
(in particular, to the case of multi-class forecasting).
%As usual, we consider the linear protocol;
Forecaster's move space is assumed to be $\mathbf{F}=\co\mathbf{Y}$.
%$\co\mathbf{Y}\subseteq\mathbf{F}\subseteq\mathbf{L}$

We say that the forecasts $f_n$ are \emph{properly calibrated}
if, for any $f^*\in\mathbf{F}$,
\begin{equation*}
  \frac
  {\sum_{i=1,\ldots,n:f_i\approx f^*} y_i}
  {\sum_{i=1,\ldots,n:f_i\approx f^*} 1}
  \approx
  f^*
\end{equation*}
provided $\sum_{i=1,\ldots,n:f_i\approx f^*} 1$ is not too small.
(We shorten $(1/c)v$ to $v/c$,
where $v$ is a vector and $c\ne0$ is a number.)
Proper calibration is only a necessary but far from sufficient condition
for good forecasts:
for example, a forecaster who ignores the data $x_n$
can be perfectly calibrated,
no matter how much useful information $x_n$ contain.
(Cf.\ the discussion in \cite{dawid:1986}.)

We say that the forecasts $f_n$ are \emph{properly calibrated and resolved}
if, for any $(f^*,x^*)\in\mathbf{F}\times\mathbf{X}$,
\begin{equation}\label{eq:prop}
  \frac
  {\sum_{i=1,\ldots,n:(f_i,x_i)\approx(f^*,x^*)} y_i}
  {\sum_{i=1,\ldots,n:(f_i,x_i)\approx(f^*,x^*)} 1}
  \approx
  f^*
\end{equation}
provided $\sum_{i=1,\ldots,n:(f_i,x_i)\approx(f^*,x^*)} 1$ is not too small.

Instead of ``crisp'' points $(f^*,x^*)\in\mathbf{F}\times\mathbf{X}$
one may consider ``fuzzy points''
$I:\mathbf{F}\times\mathbf{X}\to[0,1]$ such that $I(f^*,x^*)=1$
and $I(f,x)=0$ for all $(f,x)$
outside a small neighborhood of $(f^*,x^*)$.
A standard choice would be something like $I:=\III_{E}$,
where $E\subseteq\mathbf{F}\times\mathbf{X}$ is a small neighborhood of $(f^*,x^*)$
and $\III_{E}$ is its indicator function,
but we will want $I$ to be continuous
(it can, however, be arbitrarily close to $\III_{E}$).

Suppose $\mathbf{F}\subseteq\bbbr^m$ and $\mathbf{X}\subseteq\bbbr^l$
for some $m,l\in\{1,2,\ldots\}$.
Let $(f^*,x^*)$ be a point in $\mathbf{F}\times\mathbf{X}$;
consider a small box
$E:=\prod_{i=1}^m [a_i,b_i] \times \prod_{j=1}^l [c_j,d_j]$
containing this point, $E\ni(f^*,x^*)$.
The indicator $\III_E$ of $E$ can be arbitrarily well approximated
by the tensor product
\begin{equation*}
  I(f_1,\ldots,f_m,x_1,\ldots,x_l)
  =
  \prod_{i=1}^m F_i(f_i) \prod_{j=1}^l G_j(x_j)
\end{equation*}
of some functions $F_i$ and $G_j$ from the Sobolev class (\ref{eq:Sobolev}).
Let $\left\|I\right\|_{\FFF}$ be the norm of $I$
in the tensor product $\FFF$ of $m+l$ copies of $\SSS$
(see \cite{aronszajn:1950}, \S I.8,
for an explicit description of tensor products of RKHS).
We can rewrite (\ref{eq:calibration-resolution}) as
\begin{equation}\label{eq:variant}
  \left\|
    \frac
    {
      \sum_{i=1}^n
      I(f_i,x_i)
      (y_i-f_i)
    }
    {
      \sum_{i=1}^n
      I(f_i,x_i)
    }
  \right\|_{\mathbf{L}}
  \le
  2^{-\frac{m+l}{2}}
  \frac
  {
    \diam(\mathbf{Y})
%   \CCC_{\FFF}
    \left\|
      I
    \right\|_{\FFF}
    \sqrt{n}
  }
  {
    \sum_{i=1}^n
    I(f_i,x_i)
  }
\end{equation}
(assuming the denominator $\sum_{i=1}^n I(f_i,x_i)$
is positive);
therefore,
we can expect proper calibration and resolution
in the soft neighborhood $I$ of $(f^*,x^*)$ when
\begin{equation}\label{eq:denominator}
  \sum_{i=1}^n
  I(f_i,x_i)
  \gg
  \sqrt{n}\ifCONF\;\fi.
\end{equation}

\section{Further \UC{r}esearch}
\label{sec:further}

The main result of this paper is an existence theorem:
we did not show how to compute Forecaster's strategy
ensuring $\K_0\ge\K_1\ge\cdots$.
(The latter was easy in the case of binary forecasting
considered in \GTPVIII.)
It is important to develop
computationally efficient ways to find zeros of vector fields,
at least when $\mathbf{L}=\bbbr^m$.
There are several popular methods for finding zeros,
such as the Newton-Raphson method
(see, e.g., \cite{press/etal:1992}, Chap.~9),
but it would be ideal to have efficient methods
that are guaranteed to find a zero (or a near zero)
in a prespecified time.

\subsection*{Acknowledgments}

This work was partially supported by
% BBSRC (grant 111/BIO14428),
% EPSRC (grant GR/R46670/01),
MRC (grant S505/65),
Royal Society,
and the Superrobust Computation Project
(Graduate School of Information Science and Technology,
University of Tokyo).
We are grateful to
% COLT'2005 and ALT'2005
anonymous reviewers for their comments.

\appendix
\section{Zeros of \UC{v}ector \UC{f}ields}
% \addcontentsline{toc}{section}{Appendix A: Zeros of vector fields}
\label{app:meta-theorem}

The following lemma is the main component of the proof
of Theorem \ref{thm:main}.
\begin{lemma}\label{lem:fixed-point}
  Let $\mathbf{F}$ be a compact convex non-empty set in a Hilbert space $\mathbf{L}$
  and $S: \mathbf{F}\to\mathbf{L}$ be a continuous vector field on $\mathbf{F}$.
  If at no point of the boundary $\partial\mathbf{F}$
  the vector field $S$ is normal and directed exteriorly to $\mathbf{F}$
  then there exists $f\in\mathbf{F}$ such that $S(f)=0$.
\end{lemma}
\begin{proof}
  For each $f\in\mathbf{L}$ define $\sigma(f)$
  to be the point of $\mathbf{F}$ closest to $f$.
  A standard argument (see, e.g., \cite{rudin:1991}, Theorem 12.3)
  shows that such a point exists:
  if $d:=\inf\{\left\|y-f\right\|_{\mathbf{L}}\st y\in\mathbf{F}\}$,
  we can take any sequence $y_n\in\mathbf{F}$ with $\left\|y_n-f\right\|_{\mathbf{L}}\to d$
  and apply the parallelogram law
  $\left\|a-b\right\|^2+\left\|a+b\right\|^2=2\left\|a\right\|^2+2\left\|b\right\|^2$
  to obtain
  \begin{align*}
    \left\|
      y_m-y_n
    \right\|_{\mathbf{L}}^2
    &=
    \left\|
      (y_m-f) - (y_n-f)
    \right\|_{\mathbf{L}}^2\\
    &=
    2
    \left\|
      y_m-f
    \right\|_{\mathbf{L}}^2
    +
    2
    \left\|
      y_n-f
    \right\|_{\mathbf{L}}^2
    -
    \left\|
      (y_m-f) + (y_n-f)
    \right\|_{\mathbf{L}}^2\\
    &=
    2
    \left\|
      y_m-f
    \right\|_{\mathbf{L}}^2
    +
    2
    \left\|
      y_n-f
    \right\|_{\mathbf{L}}^2
    -
    4
    \left\|
      \frac{y_m+y_n}{2} - f
    \right\|_{\mathbf{L}}^2\\
    &\le
    2
    \left\|
      y_m-f
    \right\|_{\mathbf{L}}^2
    +
    2
    \left\|
      y_n-f
    \right\|_{\mathbf{L}}^2
    -
    4 d^2
    \to
    2 d^2 + 2 d^2 - 4 d^2
    =
    0
  \end{align*}
  as $m,n\to\infty$;
  since $\mathbf{L}$ is complete and $\mathbf{F}$ is closed,
  $y_n\to y$ for some $y\in\mathbf{F}$,
  and it is clear that $\left\|y-f\right\|_{\mathbf{L}}=d$.
  A closest point is indeed unique:
  if $\left\|y_1-f\right\|_{\mathbf{L}}=\left\|y_2-f\right\|_{\mathbf{L}}=d$
  and $y_1\ne y_2$,
  the parallelogram law would give
  \begin{multline}\label{eq:contradiction}
    \left\|
      \frac{y_1+y_2}{2} - f
    \right\|_{\mathbf{L}}^2
    =
    \frac14
    \left\|
      (y_1-f) + (y_2-f)
    \right\|_{\mathbf{L}}^2\\
    =
    \frac12
    \left\|
      y_1-f
    \right\|_{\mathbf{L}}^2
    +
    \frac12
    \left\|
      y_2-f
    \right\|_{\mathbf{L}}^2
    -
    \frac14
    \left\|
      (y_1-f) - (y_2-f)
    \right\|_{\mathbf{L}}^2\\
    =
    d^2
    -
    \frac14
    \left\|
      y_1-y_2
    \right\|_{\mathbf{L}}^2
    <
    d^2.
  \end{multline}
  Therefore,
  the function $\sigma(f)$ is well-defined.
  It is also continuous:
  if $\left\|f-\sigma(f)\right\|_{\mathbf{L}}=d$ and $f_n\to f$,
  then $\left\|f-\sigma(f_n)\right\|_{\mathbf{L}}\to d$ and,
  analogously to (\ref{eq:contradiction}),
  \begin{multline*}
    d^2
    \le
    \left\|
      \frac{\sigma(f)+\sigma(f_n)}{2} - f
    \right\|_{\mathbf{L}}^2
    =
    \frac14
    \left\|
      (\sigma(f)-f) + (\sigma(f_n)-f)
    \right\|_{\mathbf{L}}^2\\
    =
    \frac12
    \left\|
      \sigma(f)-f
    \right\|_{\mathbf{L}}^2
    +
    \frac12
    \left\|
      \sigma(f_n)-f
    \right\|_{\mathbf{L}}^2
    -
    \frac14
    \left\|
      (\sigma(f)-f) - (\sigma(f_n)-f)
    \right\|_{\mathbf{L}}^2\\
    =
    d^2 + o(1)
    -
    \frac14
    \left\|
      \sigma(f)-\sigma(f_n)
    \right\|_{\mathbf{L}}^2;
  \end{multline*}
  therefore, $\sigma(f_n)\to\sigma(f)$ in $\mathbf{L}$.

  For each $f\in\mathbf{F}$,
  let $\Sigma(f):=\sigma(f+S(f))$
  be the point of $\mathbf{F}$ closest to $f+S(f)$;
  since both $\sigma$ and $S$ are continuous,
  $\Sigma$ is continuous.
  By the Schauder-Tikhonov theorem
  (see, e.g., \cite{rudin:1991}, Theorem 5.28)
  % Schauder's fixed point theorem:
  % also \cite{agarwal/etal:2001}, Chap.~4,
  % \cite{dunford/schwartz:1958}, \S V.10
  there is a point $f\in\mathbf{F}$ such that $\Sigma(f)=f$.
  If $f$ is an interior point of $\mathbf{F}$,
  $\sigma(f+S(f))=f$ implies $S(f)=0$,
  and so the conclusion of the lemma holds.
  It remains to consider the case $f\in\partial\mathbf{F}$;
  in fact,
  we will show that this case is impossible.
  There exists $y\in\mathbf{F}$ such that
  $\langle S(f),y-f\rangle_{\mathbf{L}}>0$
  (otherwise,
  $S$ would have been normal and directed exteriorly to $\mathbf{F}$),
  and we find for $t\in(0,1)$:
  \begin{multline*}
    \left\|
      (f+S(f))
      -
      ((1-t)f + ty)
    \right\|^2_{\mathbf{L}}
    =
    \left\|
      S(f)
      -
      t(y-f)
    \right\|^2_{\mathbf{L}}\\
    =
    \left\|
      S(f)
    \right\|^2_{\mathbf{L}}
    -
    2t
    \left\langle
      S(f),
      y-f
    \right\rangle_{\mathbf{L}}
    +
    t^2
    \left\|
      y-f
    \right\|^2_{\mathbf{L}};
  \end{multline*}
  for a small enough $t$ this gives
  \begin{equation*}
    \left\|
      (f+S(f))
      -
      ((1-t)f + ty)
    \right\|^2_{\mathbf{L}}
    <
    \left\|
      S(f)
    \right\|^2_{\mathbf{L}},
  \end{equation*}
  a contradiction.
  \qedtext
\end{proof}

\ifFULL
\subsection*{The \UC{c}ase $\dim(\mathbf{L})=2$}

In this subsection we give a relatively efficient way
of finding a zero of $S_n$
for the case $\dim(\mathbf{L})=2$
(following Aleksandrov \cite{aleksandrov:1969}, \S5, p.~217).
In this subsection ``region'' means a connected open subset of $\mathbf{L}$
with a piecewise smooth boundary.

Consider a region $R$ containing $\mathbf{Y}$.
For each closed curve $C$ in $R$
we define its \emph{winding number} $w(C)$ as follows:
when we make a full circle around the curve in the positive direction,
the angle that the direction of the field $S_n$
forms with an arbitrarily fixed direction
returns to its initial value increased
as result of making the full circle
by $2\pi w(C)$.
The winding number of a region in $R$ is the winding number of its boundary.
The winding number of $R$ itself is non-zero
since there are zeros of $S_n$ inside $R$.
Divide $R$ into two regions, find their winding numbers,
and discard one of the regions so that the remaining region
has a non-zero winding number.
Again divide the remaining region into two regions, etc.
\fi

\section{Tensor \UC{p}roduct}
% \addcontentsline{toc}{section}{Appendix B: Tensor product}
\label{app:tensor}

In this appendix we list several definitions and simple facts
about tensor products of Hilbert spaces,
in the form used in this paper.

The \emph{tensor product} $\mathbf{L}\otimes\mathbf{H}$
of Hilbert spaces $\mathbf{L}$ and $\mathbf{H}$
is defined in, e.g., \cite{reed/simon:1972}, \S II.4.
Briefly, the definition is as follows.
The space $\mathbf{L}\otimes\mathbf{H}$
is the subset of the set of bilinear forms $v(l',h')$,
$l'\in\mathbf{L}$ and $h'\in\mathbf{H}$,
obtained as the completion of the set of all linear combinations
of the bilinear forms $l\otimes h$, where $l\in\mathbf{L}$ and $h\in\mathbf{H}$,
defined by
\begin{equation}\label{eq:def1}
  (l\otimes h)(l',h')
  :=
  \langle
    l,l'
  \rangle_{\mathbf{L}}
  \langle
    h,h'
  \rangle_{\mathbf{H}};
\end{equation}
the inner product in $\mathbf{L}\otimes\mathbf{H}$ is determined uniquely
by setting
\begin{equation}\label{eq:def2}
  \left\langle
    l_1\otimes h_1,
    l_2\otimes h_2
  \right\rangle_{\mathbf{L}\otimes\mathbf{H}}
  :=
  \left\langle
    l_1, l_2
  \right\rangle_{\mathbf{L}}
  \left\langle
    h_1, h_2
  \right\rangle_{\mathbf{H}}
  \ifCONF\;\fi.
\end{equation}
% The only properties of this construction we need in this paper
% are stated in the following lemmas.
In particular,
(\ref{eq:def2}) implies
\begin{equation}\label{eq:implication}
  \left\|
    l\otimes h
  \right\|_{\mathbf{L}\otimes\mathbf{H}}
  =
  \left\|l\right\|_{\mathbf{L}}
  \left\|h\right\|_{\mathbf{H}}
\end{equation}
for all $l\in\mathbf{L}$ and $h\in\mathbf{H}$.

If $v\in\mathbf{L}\otimes\mathbf{H}$ and $h\in\mathbf{H}$,
we define the \emph{product} $vh\in\mathbf{L}$ by the requirement
\begin{equation*}
  v(l',h)
  =
  \left\langle
    vh,l'
  \right\rangle_{\mathbf{L}},
  \enspace
  \forall l'\in\mathbf{L}
\end{equation*}
(the validity of this definition follows from the Riesz-Fischer theorem:
all bilinear forms in $\mathbf{L}\otimes\mathbf{H}$ are clearly continuous).
\begin{lemma}\label{lem:tensor3}
  For any $l\in\mathbf{L}$ and $h_1,h_2\in\mathbf{H}$,
  \begin{equation}\label{eq:to-prove}
    (l\otimes h_1) h_2
    =
    \langle
      h_1, h_2
    \rangle_{\mathbf{H}}
    l
    \ifCONF\;\fi.
  \end{equation}
\end{lemma}
\begin{proof}
  It suffices to prove
  \begin{equation*}
    \left\langle
      (l\otimes h_1) h_2,
      l'
    \right\rangle_{\mathbf{L}}
    =
    \langle
      h_1, h_2
    \rangle_{\mathbf{H}}
    \langle
      l, l'
    \rangle_{\mathbf{L}}
    \ifCONF\;\fi,
  \end{equation*}
  which, by definition, is equivalent to
  \begin{equation*}
    (l\otimes h_1)
    (l',h_2)
    =
    \langle
      h_1, h_2
    \rangle_{\mathbf{H}}
    \langle
      l, l'
    \rangle_{\mathbf{L}}
  \end{equation*}
  and, therefore, true
  (cf.\ (\ref{eq:def1})).
  \qedtext
\end{proof}
The following lemma is an easy implication of the Cauchy-Schwarz inequality.
\begin{lemma}\label{lem:tensor2}
  For any $v\in\mathbf{L}\otimes\mathbf{H}$ and $h\in\mathbf{H}$,
  \begin{equation*}
    \|vh\|_{\mathbf{L}}
    \le
    \|v\|_{\mathbf{L}\otimes\mathbf{H}}
    \|h\|_{\mathbf{H}}
    \ifCONF\;\fi.
  \end{equation*}
\end{lemma}
\begin{proof}
  We are required to prove, for all $l'\in\mathbf{L}$,
  \begin{equation*}
    \left\langle
      vh,
      l'
    \right\rangle_{\mathbf{L}}
    \le
    \|v\|_{\mathbf{L}\otimes\mathbf{H}}
    \|h\|_{\mathbf{H}}
    \|l'\|_{\mathbf{L}}
    \ifCONF\;\fi,
  \end{equation*}
  i.e.,
  \begin{equation*}
    v(l',h)
    \le
    \|v\|_{\mathbf{L}\otimes\mathbf{H}}
    \|h\|_{\mathbf{H}}
    \|l'\|_{\mathbf{L}}
    \ifCONF\;\fi.
  \end{equation*}
  We can assume that $v=l\otimes h'$,
  for some $l\in\mathbf{L}$ and $h'\in\mathbf{H}$,
  in which case the last inequality immediately follows
  from (\ref{eq:def1}), (\ref{eq:implication}), and the Cauchy-Schwarz inequality.
  \qedtext
\end{proof}

%\ifWP
%\DFlastpage
%\fi
\end{document}